\documentclass[10pt,twocolumn,letterpaper]{article}

\usepackage{cvpr}              % To produce the CAMERA-READY version
\usepackage{graphicx}
\usepackage{amsmath}
\usepackage{amssymb}
\usepackage{booktabs}

\usepackage{url}
\usepackage{epsfig}
\usepackage{graphicx}
\usepackage{multirow}
\usepackage{multirow}
\usepackage{caption}
\usepackage[ruled]{algorithm2e}
\usepackage{bbding}

\usepackage{subcaption}
\usepackage{makecell}
\usepackage[table]{xcolor}

\usepackage{amsmath,amsfonts,bm}

\def\eqref#1{equation~\ref{#1}}
\def\1{\bm{1}}

\DeclareMathAlphabet{\mathsfit}{\encodingdefault}{\sfdefault}{m}{sl}
\SetMathAlphabet{\mathsfit}{bold}{\encodingdefault}{\sfdefault}{bx}{n}

\usepackage[pagebackref,breaklinks,colorlinks]{hyperref}

\usepackage[accsupp]{axessibility}  % Improves PDF readability for those with disabilities.

\usepackage[capitalize]{cleveref}
\crefname{section}{Sec.}{Secs.}
\Crefname{section}{Section}{Sections}
\Crefname{table}{Table}{Tables}
\crefname{table}{Tab.}{Tabs.}

\def\cvprPaperID{5534} % *** Enter the CVPR Paper ID here
\def\confName{CVPR}
\def\confYear{2022}

\begin{document}

\title{Training-free Transformer Architecture Search}

\author{Qinqin Zhou$^{1}$$^{*}$ \quad\quad\quad Kekai Sheng$^{2}$\thanks{The first two authors contributed equally. This work was done when Qinqin Zhou  was intern at Tencent Youtu Lab.}\quad\quad\quad Xiawu Zheng$^{3}$\quad\quad\quad Ke Li $^{2}$ \\Xing Sun$^{2}$\quad\quad\quad Yonghong Tian$^{4}$\quad\quad\quad Jie Chen$^{4,3}$\quad\quad\quad Rongrong Ji$^{1,3,5,6}$\thanks{Corresponding author: rrji@xmu.edu.cn.} \\$^1$Media Analytics and Computing Lab,  School of Informatics, Xiamen University,
$^2$Tencent Youtu Lab,
\\$^3$Peng Cheng Laboratory,
$^4$School of Electronic and Computer Engineering, Peking University,
\\$^5$Institute of Artificial Intelligence, Xiamen University,
$^6$Fujian Engineering Research \\Center of Trusted Artificial Intelligence Analysis and Application, Xiamen University
}
\maketitle

%%%%%%%%% ABSTRACT
\begin{abstract}
%%% 2021/11/15
Recently, Vision Transformer (ViT) has achieved remarkable success in several computer vision tasks. The progresses are highly relevant to the architecture design, then it is worthwhile to propose Transformer Architecture Search (TAS) to search for better ViTs automatically.
However, current TAS methods are time-consuming and existing zero-cost proxies in CNN do not generalize well to the ViT search space according to our experimental observations.
In this paper, for the first time, we investigate how to conduct TAS in a training-free manner and devise an effective training-free TAS (TF-TAS) scheme. Firstly, we observe that the properties of multi-head self-attention (MSA) and multi-layer perceptron (MLP) in ViTs are quite different and that the synaptic diversity of MSA affects the performance notably. Secondly, based on the observation, we devise a modular strategy in TF-TAS that evaluates and ranks ViT architectures from two theoretical perspectives: synaptic diversity and synaptic saliency, termed as DSS-indicator. With DSS-indicator, evaluation results are strongly correlated with the test accuracies of ViT models.
Experimental results demonstrate that our TF-TAS achieves a competitive performance against the state-of-the-art manually or automatically design ViT architectures, and it promotes the searching efficiency in ViT search space greatly: from about $24$ GPU days to less than $0.5$ GPU days.
Moreover, the proposed DSS-indicator outperforms the existing cutting-edge zero-cost approaches (\emph{e.g.,} TE-score and NASWOT). 
\end{abstract}

\begin{figure}
    \centering
    \includegraphics[width=0.98\linewidth,height=7.5cm]{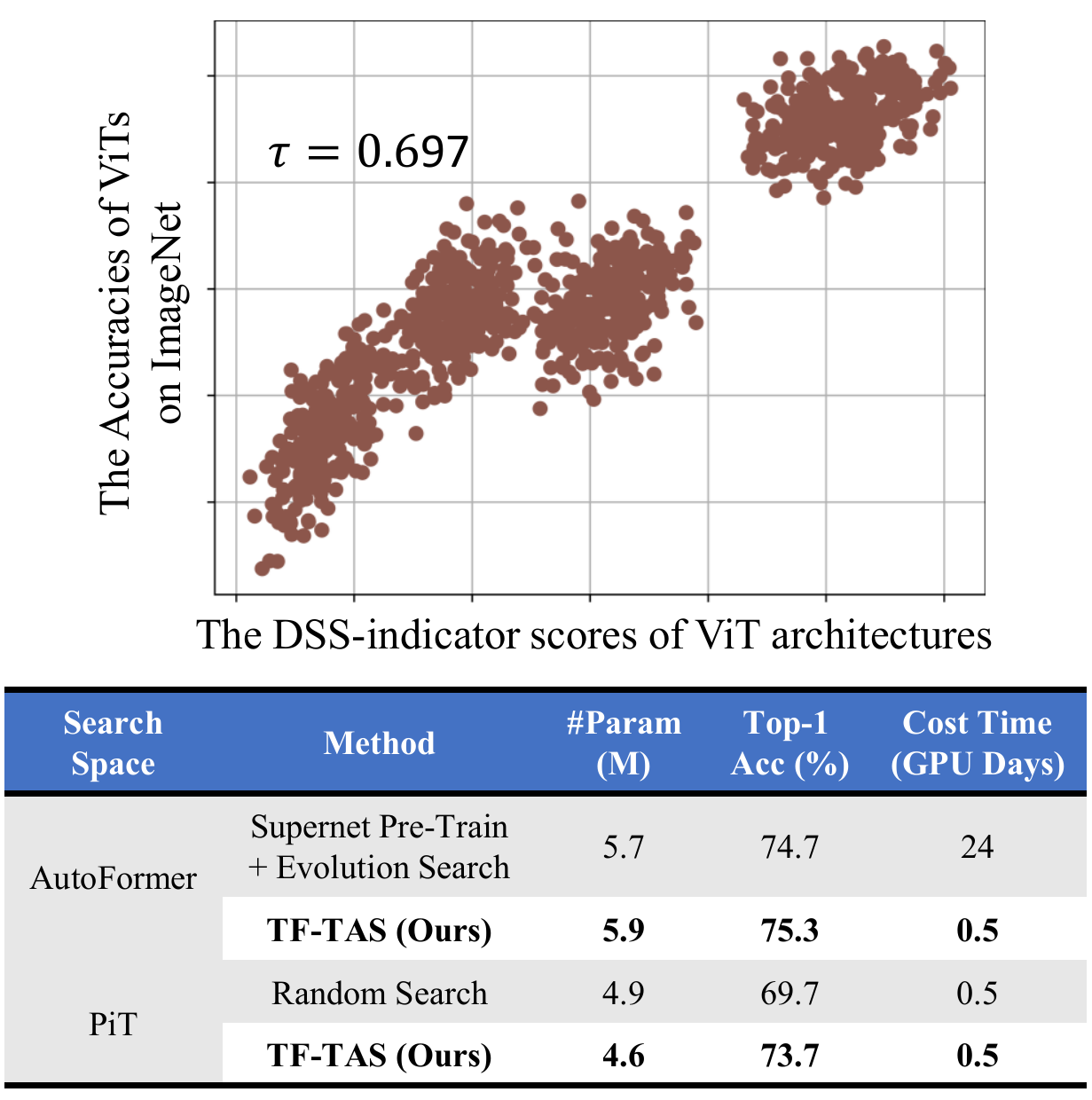}
    \caption{\textbf{Top}: The evaluation results from DSS-indicator are positively correlated with the test accuracies of various ViT networks.
    \textbf{Bottom}: The design and computation cost of existing methods, \ie, AutoFormer~\cite{7} and PiT~\cite{17}, and the proposed TF-TAS.
    }
    \label{fig:teaser}
    \vspace{-3mm}
\end{figure}

%%%%%%%%% BODY TEXT
\section{Introduction}
Vision Transformer (ViT)~\cite{1,33,hu2021istr} has shown its competitiveness in the computer vision community recently, and has been an important research hot-spot.
With the emergence of manually-designed advanced ViT models~\cite{2,3,4}, Transformer Architecture Search (TAS)~\cite{5,6,7,8,28} makes its grand debut and aims at searching multiple configurations of ViT architecture in an automated way. 
Although the one-shot NAS scheme~\cite{9,47,zheng2021migo,zheng2021evolving} is leveraged in TAS, it still demands a high computational cost (\eg, larger than $24$ GPU days) to train a supernet for reliable performance estimations on various ViT architectures.
Furthermore, since the magnitude of ViT search spaces (\eg, $\sim 10^{30}$ in GLiT~\cite{28}) far exceeds the magnitude of CNN search spaces (\eg, $\sim 10^{18}$ in DARTS~\cite{liu2018darts}) and ViT models usually require more training epochs (\emph{e.g.}, $300$), the search efficiency of one-shot based TAS is still unsatisfying.

Recall that, to promote the searching efficiency on CNN search spaces, several proxies (\emph{e.g.}, GraSP~\cite{11}, TE-score~\cite{16}, and NASWOT~\cite{15}) are proposed to evaluate the rank of different CNN architectures in a zero-cost manner.
Technically, one typical CNN is mainly composed of convolution layers. On the other hand, the basic blocks of one ViT model, multi-head self-attention (MSA) and multi-layer perceptron (MLP), are mainly composed of linear layers. The difference makes it risky to apply existing zero-cost proxies that are verified on CNN directly to the ViT search space.
Consequently, it is necessary and worthwhile to investigate the possibility of an effective zero-cost proxy that is more suitable for ranking ViT networks and facilitates the training efficiency of TAS.
This problem motivates us to delve further into the ViT architecture to propose an effective approach to conduct TAS in a training-free manner.

To this end, we conduct a modular investigation on MSA and MLP in typical ViTs~\cite{7,17} to devise effective performance indicator methods for MSA and MLP in ranking ViT networks.
Based on numerical results, we observe that the MSA and MLP in ViT have different properties in indicating the performance of the model. When the MSA possesses higher \textit{diversity} score or when the MLP has more \textit{synaptic saliency} value, the corresponding ViT network always has better performance (see the \textit{Top} of \cref{fig:teaser}).
Motivated by the significant insights, we propose an effective and efficient DSS-indicator and design a training-free TAS (TF-TAS). 
Specifically, we attempt to comprehensively rank various ViTs by exploiting the properties of the MSA and MLP mentioned above. DSS-indicator estimates the synaptic diversity of MSA and the synaptic saliency of MLP to generate effective evaluation scores for ViT architectures. The synaptic diversity measures the degree of rank collapse on one MSA, and the synaptic saliency estimates the amount of important parameter within one MLP.
To the best of our knowledge, it is the first time to propose synaptic diversity of MSA and synaptic saliency of MLP as the proxy in evaluating ViT architectures. Besides, it should be noted that our TF-TAS is orthogonal to the search space design and weight sharing strategy. Thus, it is flexible to combine TF-TAS with other ViT search space or TAS methods to further promote the searching efficiency.
Compared with the manually designed ViTs~\cite{33,36,TNT,SwinTransformer} and the automatically searched ones~\cite{7,8,28}, our TF-TAS achieves a competitive performance and accelerates the searching procedure from around $24$ GPU days to less than $0.5$ GPU days, about $48$ times faster (see the \textit{Bottom} of \cref{fig:teaser}).

For fair comparison and full investigation, we also present a reliable test-bed to evaluate the \textit{state-of-the-art} zero-cost proxies (\eg, TE-score~\cite{16} and NASWOT~\cite{15}) in the ViT search space.
We construct a large proxy ViT benchmark, which is based on several pre-trained supernets of AutoFormer~\cite{7}, to compare the relative performance of alternative zero-cost proxies on ViT architectures.
With the numerical observations of zero-cost proxies, we empirically verify the relative rankings of different zero-cost proxies in TAS, and our DSS-indicator outperforms the other counterparts. We also draw some practical insights in designing a better proxy for ranking ViT architectures.

Overall, our principal contributions are listed as follows:
\begin{itemize}
  \item We propose a training-free TAS (TF-TAS), which includes a modular strategy to combine the synaptic diversity of the MSAs and synaptic saliency of the MLPs as a DSS-indicator in evaluating ViT architectures.

  \item Extensive experiments demonstrate that the proposed TF-TAS not only achieves a competitive search performance, but also improves the search efficiency in searching ViT architectures.
  
  \item We design a series of controlled experiments to compare the existing zero-cost proxies in TAS. The results provide some empirical insights in designing optimal proxy metric in evaluating ViT architectures.
\end{itemize}

\section{Preliminary}
\label{Preliminary}

\paragraph{Design \& Search Transformer Architectures.}
Since ViT~\cite{1}, the computer vision community has witnessed the emergence of many manually designed advanced ViT architectures~\cite{2,3,17,49}. Technically, most of them consist of the same basic blocks that include MSA, Layer Normalization (LN), and MLP. Existing TAS approaches~\cite{8,28,7} search different dimensions in MSAs and MLPs, such as the number of heads in MSAs, the ratio of MSAs or MLPs. These methods are generally built on the one-shot NAS framework~\cite{9,47,zheng2021migo,zheng2021evolving}: train the supernet by training a subnet path in each epoch.
The typical one-shot based TAS method AutoFormer~\cite{7} includes an entanglement strategy and divides the search space into three sub-supernets, each of which are trained in a one-shot manner for $500$ epochs (about $24$ GPU days) and requires $8$ NVIDIA V100 GPUs. The size of AutoFormer search space is $1.7 \times 10^{16}$, which makes it time-consuming in training supernet.

In general, how to reduce the cost of searching the ViT architectures and ensuring the performance of the searched networks is a fundamental and challenging problem.
In this paper, we try to find a way to maintain the performance of TAS and accelerate the searching efficiency.

\paragraph{Performance Evaluation via Zero-cost Proxy.}
There are two mainstreams of zero-cost proxy to reduce the cost of performance estimation and promote searching efficiency.
The first one, inspired by the pruning community, sums up the saliency value of each model weight as the proxy of the corresponding CNN architecture with a single forward / backward propagation. The popular methods include Grad-norm~\cite{14}, SNIP~\cite{10}, and GraSP~\cite{11}. 
These proxies follow a default assumption: the more salient the weight value is, the more important it is to the model; and the more salient weight one network has, the better performance the model does. 
The second one, such as TE-score~\cite{16}, NASWOT~\cite{15}, and Zen-Score~\cite{31}, is designed specifically for CNNs.
They analyze the important properties (\emph{e.g.}, expressivity) of the representations of CNNs. 
Mellor~\etal~\cite{15} proposed jacobian covariance to sum up the saliency of each weight 
to rank CNNs. Chen~\etal~\cite{16} applied two theory-inspired indicators as the proxy to find the best subnet. 

Different from the existing literature, in this paper, for the first time, we identify the shortcoming of directly applying existing proxies in the ViT search space. Then, we propose a simple yet effective proxy to generate better performance and facilitate the searching efficiency of TAS.

\section{Methodology}
\label{sec:Methodology}
\subsection{Motivation}
The existing TAS methods~\cite{8,28,7} are relatively time-consuming, especially in performance estimation (\eg, 300 training epochs on $8$ GPUs). Then, it is worthwhile to leverage the zero-cost proxies~\cite{11,15,16} to rank ViT architectures and reduce the computation cost in performance estimation. Nevertheless, the existing zero-cost proxies are specifically designed for the CNN search spaces (\eg, DARTS~\cite{liu2018darts} and NAS-Bench 201~\cite{dong2019NASBench201}). Obviously, the search space of ViT is quite different from that of CNN, then the existing proxies could not promise generalization on the ViT search space (see the results in~\cref{sec:VS_other_zero-cost}). It motivates us to explore and exploit the useful properties of MSA and MLP within ViT, and design an effective ViT-oriented zero-cost proxy.

In this section, \textcolor{black}{we first propose an effective method to calculate the synaptic diversity of MSA and generate evaluation results that are positively correlated with the classification accuracies of ViT networks.} 
\textcolor{black}{Then, we find that when the MLP module has more important weight parameters, \ie, higher synaptic saliency value, the corresponding ViT network always yields better classification performance.}
Finally, we propose a \textcolor{black}{DSS-indicator} 
to comprehensively exploit the properties of MSAs and MLPs in ranking various ViT architectures effectively and efficiently, which promotes the searching efficiency of TAS.

\begin{figure*}
    \centering
    \begin{subfigure}[b]{0.33\textwidth}
         \centering
         \includegraphics[width=\textwidth,height=4.4cm]{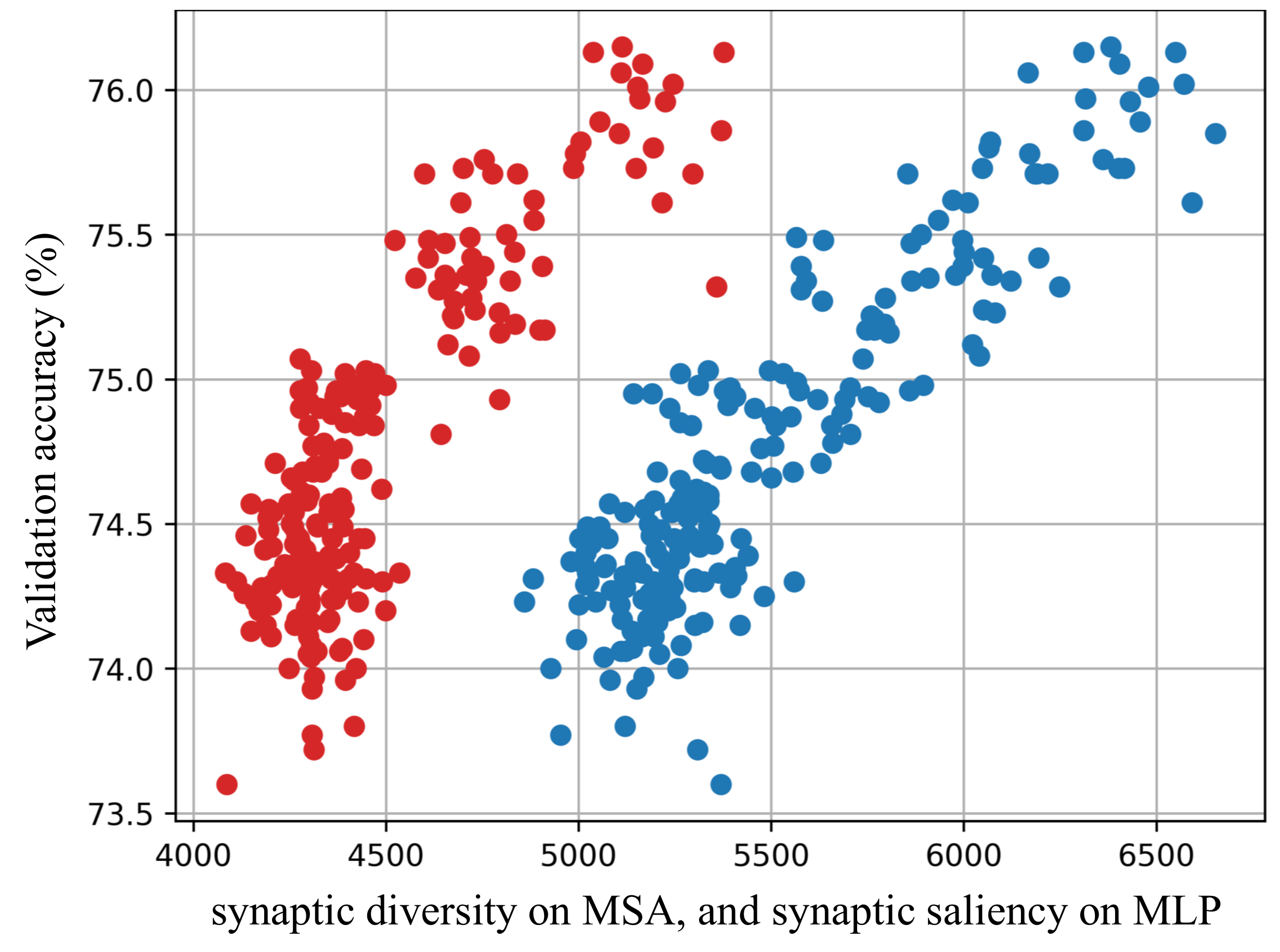}
         \caption{}
         \label{fig:MSAMLP_testAccuracy}
    \end{subfigure}
    % \hfill
    \begin{subfigure}[b]{0.32\textwidth}
         \centering
         \includegraphics[width=\textwidth,height=4.4cm]{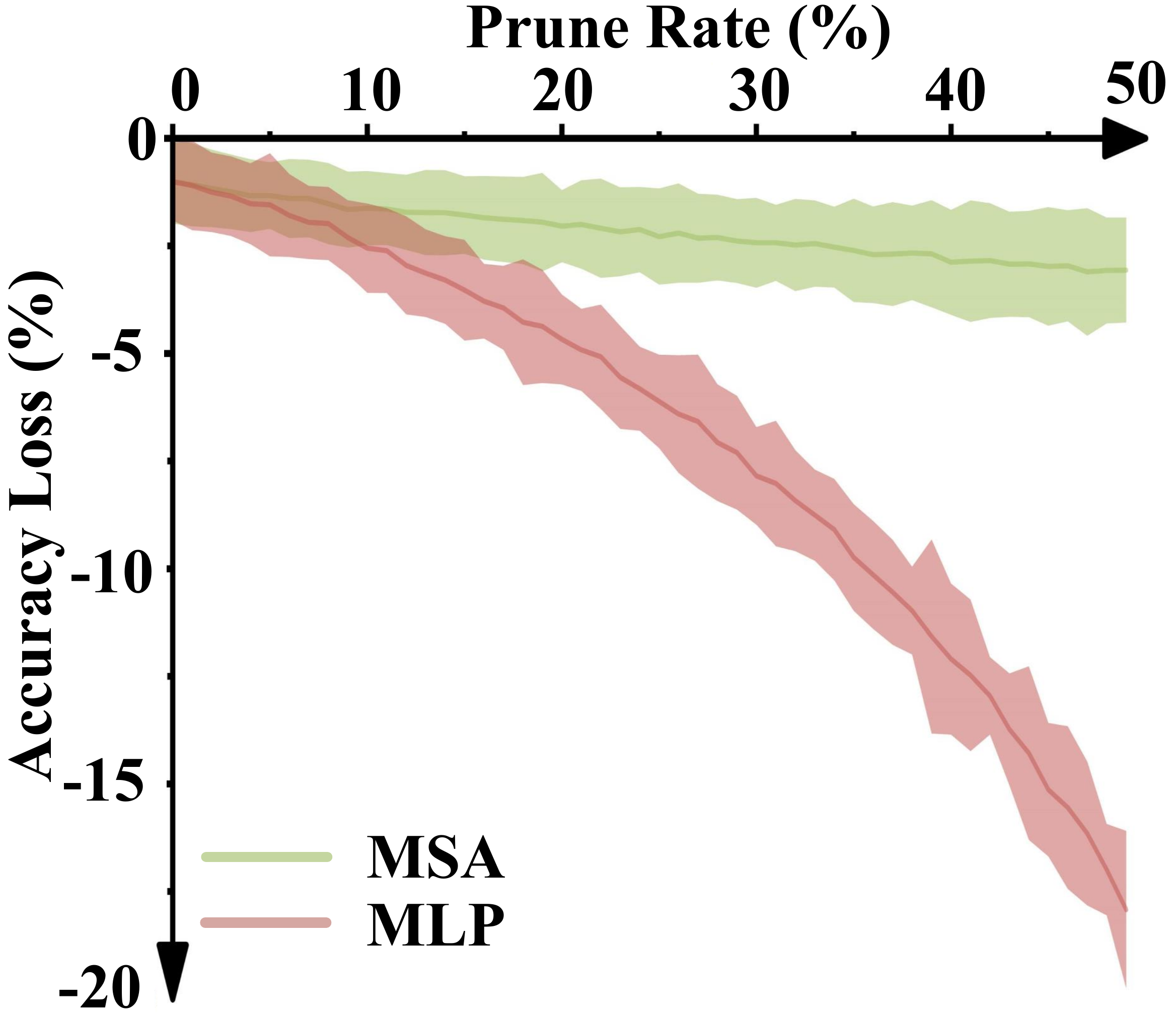}
        \caption{}
         \label{fig:prune_PT}
    \end{subfigure}
    % \hfill
    \begin{subfigure}[b]{0.32\textwidth}
         \centering
         \includegraphics[width=\textwidth,height=4.4cm]{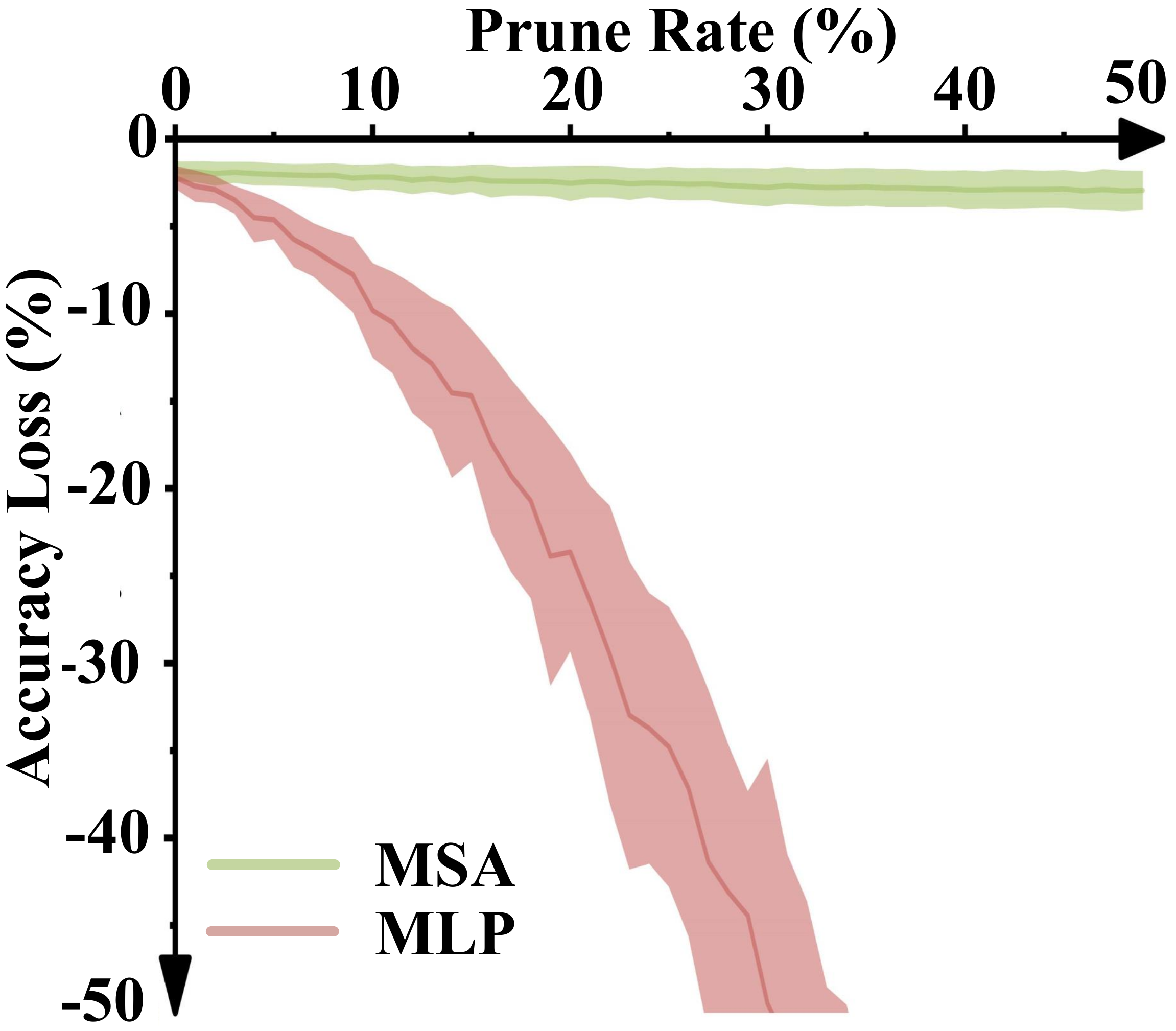}
        \caption{}
         \label{fig:prune_HT}
    \end{subfigure}
    \caption{(\textbf{a}) Illustrations of the positive connection between the proxy score ($D_{MSA}^{l}$ \textcolor{red}{red points} \& $S_{MLP}^{l}$ \textcolor{blue}{blue points}) and the accuracy of the ViT model.
    (\textbf{b}) \& (\textbf{c}) The sensitivity analysis of MSA and MLP to pruning on a flat ViT~\cite{7} and a deep-narrow ViT~\cite{17}, respectively.}
    \label{fig:SensitivityAnalysis}
    \vspace{-3mm}
\end{figure*}

\subsection{Synaptic Diversity in MSA}
\paragraph{Theoretical Analysis.}
MSA is a basic fundamental component of ViT architectures. \textcolor{black}{Several works unveil one important property of MSA: its \textit{diversity}~\cite{37,46}.} In particular, Dong~\etal~\cite{46} pointed out that the MSA causes \textit{rank collapse} in the learned representations. In specific, as the input propagates forward in the network and the depth continues to deepen, the outputs of MSAs in ViTs gradually converge to rank-1.
\textcolor{black}{And eventually, the output degenerates into a matrix with a rank of $1$, the value of each row becomes the same, \ie the scarcity of diversity. Such rank collapse severely degenerates the performance of ViT.}

However, estimating the rank collapse in the high dimension representation space requires a huge computation cost. Actually, Fazel~\etal~\cite{48} demonstrates that the rank of a matrix contains representative cues of the diversity information within the features.
Building on these understandings, the rank of the weight parameters in the MSA module could be adopted as an indicator to evaluate the ViT architecture.

\vspace{-2mm}
\paragraph{Synaptic Diversity.}
For the MSA module, it is still computationally complex to directly measure the rank of its weight matrix and hinders practical applications.
To accelerate the calculation of synaptic diversity in MSA module, we leverage the Nuclear-norm of the MSA's weight matrix to approximate its rank as the diversity indicator.
Theoretically, the Nuclear-norm of a weight matrix can be treated as an equivalent substitution for its rank, when the Frobenius-norm of the weight matrix meets certain conditions. Specifically, we denote the weight parameter matrix of an MSA module as $W_{m}$. $m$ indicates the $m$-th linear layer in an MSA module.
The Frobenius-norm of $W_{m}$ is defined as:
\begin{equation}
\begin{aligned}
\left \| W_{m} \right \|_{F}=\sqrt{\sum_{i=1}^{U}\sum_{j=1}^{V}\left | w_{i,j} \right |^{2}},
\end{aligned}\end{equation}
where $U, V$ are the dimension of $W_{m}$, and $w_{i,j}$ denotes the the element in the $i$-th row and $j$-th column of $W_{m}$.
According to inequality of arithmetic and geometric means, the upper-bound of $W_{m}$ is calculated as:
\begin{equation} \begin{split}
 \left \| W_{m} \right \|_{F}&\quad \leq \sqrt{\sum_{i=1}^{U}(\sum_{j=1}^{V} w_{i,j})\cdot (\sum_{j=1}^{V} w_{i,j})} \\ &\quad =\sqrt{\sum_{i=1}^{U}1\cdot 1}=\sqrt{U}.
\end{split}\end{equation}
This means the upper-bound of $\left \| W_{m} \right \|_{F}$ is the largest number of linear independent vectors of $ W_{m}$, \emph{i.e.}, the matrix rank.
Given two randomly selected vectors $w_{m}^{i}$ and $w_{m}^{j}$ in $W_{m}$, $\left \| W_{m} \right \|_{F}$ could be larger when $w_{m}^{i}$ and $w_{m}^{j}$ are independent. This indicates that: the larger the Frobenius-norm of $W_{m}$ goes, the closer the rank of $W_{m}$ is to the diversity of $W_{m}$. And according to one \textit{Theorem} proved by Fazel~\etal~\cite{48} that when $\left \| W_{m} \right \|_{F} \leq 1$, the Nuclear-norm of $W_{m}$ could be an approximation of the rank of $W_{m}$. Formally, the Nuclear-norm of $W_{m}$ is defined as:
\begin{equation}
\begin{aligned}
\left \| W_{m} \right \|_{nuc}= tr(\sqrt{W_{m}^{T}W_{m}}),
\end{aligned}\end{equation}
where $tr(*)$ denotes the trace of the corresponding matrix. In our case, we always have $\left \| W_{m} \right \|_{F} \leq \sqrt{U}$, thus the rank of $W_{m}$ could be approximated as $\left \| W_{m} \right \|_{nuc}/\sqrt{U}$. Theoretically, $\left \| W_{m} \right \|_{nuc}$ is proportional to $\left \| W_{m} \right \|_{nuc}/\sqrt{U}$, which indicate that the Nuclear-norm of $W_{m}$ could be used to measure the diversity of $W_{m}$. \textcolor{black}{To better estimate the synaptic diversity of MSA modules from one ViT network that the weights are randomly initialized, we further consider the aforementioned procedure on the gradient matrix $\partial \mathcal L / \partial W_{m}$ ($\mathcal{L}$ is the loss function) of each MSA module.}

Overall, we define the synaptic diversity of the weight parameter in the $l$-th MSA module as follows:
\begin{equation}
\begin{aligned}
D_{MSA}^l=\sum _{m} \left \|\frac{\partial \mathcal L}{\partial W_{m}}\right \|_{nuc}\odot \left \| W_{m} \right \|_{nuc}.
\end{aligned}
\label{q2}
\end{equation}

To verify the positive correlation between the synaptic diversity of MSA and the test accuracy of the given ViT architectures, we re-train $100$ ViT networks sampled from AutoFormer~\cite{7} and obtain their corresponding classification performance and synaptic diversity of MSA modules. The Kentall's $\tau$ between them is $0.65$ as shown in \cref{fig:MSAMLP_testAccuracy}.
The results in \cref{sec:Experiments} will also demonstrate the positive connection between the evaluation score of Eq.(\ref{q2}) and the performance for each input ViT architecture.

\subsection{Synaptic Saliency in MLP}
\label{Synaptic}
\paragraph{Theoretical Analysis.}
Network pruning~\cite{25,meng2020pruning} has achieved many progresses for CNNs, and start showing its power on Transformer~\cite{22,23,24}. Several effective CNN punning methods~\cite{12,34,35} have been proposed to measure the importance of the model weights at the early training stage. Tanaka~\etal~\cite{12} measured the synaptic saliency in pruning CNNs without training. 
Wang~\etal~\cite{22} find that different modules in Transformers show different degrees of redundancy even at the initialization stage, and tried to prune the different dimensions of Transformers. Similar to pruning, TAS focuses on searching several important dimensions, which include the number of attention heads, MSA and MLP ratio, \emph{etc}.
Inspired by these developments, we try to use the synaptic saliency to evaluate different ViTs.
However, it has been validated~\cite{22,23,24} that the sensitivities of MSA and MLP to pruning are different. A large percentage of weight in the MSA is redundant~\cite{23,24}, which has little impact on the performance at test time. 
It implies that synaptic saliency might demonstrate different performances in MSA and MLP.

To further validate the findings, we through a pruning sensitive experiment to show some quantitative results. 
As illustrated in \cref{fig:prune_PT}, we randomly sample $5$ ViT architectures from AutoFormer search space to analyze the sensitivity of the MSA and MLP to pruning.
We observe that the MLP is much more sensitive to pruning than the MSA.
We also conduct the analysis on the deep-narrow ViT networks (\eg, PiT \cite{17}), and obtain similar observations (see \cref{fig:prune_HT}). Moreover, we adopt synaptic saliency on MSA and MLP modules as proxies to calculate the Kendall's $\tau$ on the proxy ViT benchmark, respectively. The Kendall's $\tau$ of synaptic saliency on the MLP is 0.47, which is better than on the MSA (0.24) and both of the MLP and MSA (0.41). 

Since the synaptic saliency is generally calculated in the form of summation, the redundancy brings a cumulative effect. Specifically, the MSA module proves to be insensitive to pruning, which means the weight parameters of the MSAs have a higher redundancy. It has been proposed in the pruning community~\cite{14} that the values of the redundant weight parameters are much smaller than the non-redundant ones. Although the values of these redundant parameters are relatively small, redundancies of more than 50 percent tend to have a large cumulative effect, especially when distinguishing between similar architectures. For the cumulative effect, the redundant weight parameters of MSA are considered in the zero-cost proxies to measure the saliency, the cumulative forms in the zero-cost proxies lead to the cumulative effect in the MSA. The cumulative effect might make the zero-cost proxies give a higher rank to the network. 
Meanwhile, the synaptic saliency of the MLP modules is less affected by weight redundancy, which could be adopted as an indicator for the MLP modules.

\vspace{-2mm}
\paragraph{Synaptic Saliency.}
To evaluate the MLP in ViT, we resort to the \textit{synaptic saliency}.
It is extensively studied in network pruning to indicate the importance of model weights. There are several pruning-based zero-cost proxies \cite{14,10,11} that could be directly used to measure the synaptic saliency of CNNs for the CNNs are mainly composed of convolution layers. On the other hand, ViT architectures are mainly composed of MLP and MSA modules, which have different pruning properties. Through the pruning sensitivity analysis of the MSA and MLP modules in \cref{Synaptic}, we validate that the MLP module is much more sensitive to pruning. Therefore, the differences in the importance of weights in MLP modules can be better reflected by the synaptic saliency. As a comparison, the MSA modules are relatively insensitive to the punning, the synaptic saliency of which is often influenced by the redundant weights.

Building on the pruning sensitivity of MLP, we propose to measure the synaptic saliency in a modular manner. Specifically, the proposed modular strategy measures the synaptic saliency of MLPs as a part of the indicator for a ViT architecture. 
Formally, given a ViT architecture, the saliency score of the $l$-th MLP module is:
\begin{equation}
\begin{aligned}
S_{MLP}^{l}=\sum _{n} \frac{\partial \mathcal L}{\partial W_{n}}\odot W_{n}.
\end{aligned}\label{q3}\end{equation}
where $n$ denotes the number of linear layer in the $l$-th MLP in a specified ViT network, which usually set to $2$.

\cref{fig:MSAMLP_testAccuracy} shows some qualitative results to verify the effectiveness of $S_{MLP}$ in evaluating ViT architectures.

\subsection{Training-free TAS}
\label{Ourmethod}

Based on the above analysis, we propose a training-free TAS (TF-TAS) with a modular strategy to further improve search efficiency.
The modular strategy is proposed to consider the training-free evaluation of ViT architectures in two parts to form a DSS-indicator.

Combine the synaptic diversity of MSA and the saliency score of MLP, we formulate the DSS-indicator as follows:
\begin{equation}
\begin{aligned}
S_{DSS}(\mathcal A)=\sum _{l} D_{MSA}^l + \sum _{k} S_{MLP}^k.
\end{aligned}\label{q4}\end{equation}
Overall, the DSS-indicator evaluates each ViT architecture from two different perspectives.
TF-TAS calculates $S_{DSS}$ after a forward and backward
as the indicator of a specified ViT architecture. We keep each pixel of the input data being 1 to eliminate the affection of input data. 
Thus, $S_{DSS}$ is invariant to random seed. Moreover, the loss at first iteration is expressed as:
$
\mathcal{L}=\mathbb{I}^{T}\left ( \prod_{l}^{L} \left|\omega ^{[l]} \right|\right )\mathbb{I},
$
where $\mathbb{I}$ is the all ones vector. It makes Eq.(\ref{q4}) to take the inter-layer interactions of weight parameters into account to measure the diversity of MSAs and the saliency of MLPs.

\begin{table*}[ht]
\centering
\newcommand{\tabincell}[2]{\begin{tabular}{@{}#1@{}}#2\end{tabular}}
\caption{The comparison results on the Autoformer search space. $*$ denotes the results reported by \cite{17}.
}
\label{tab:MainResultsOnImageNet}
\vspace{-2mm}
\setlength{\tabcolsep}{2.4mm}
\small{
\begin{tabular}{l|cc|cc|ccc}
\toprule
% \hline 
Models       & \multicolumn{1}{c}{\tabincell{c}{\#Param (M)}} & \tabincell{c}{FLOPS (B)} & \multicolumn{1}{c}{\tabincell{c}{Top-1 (\%)}} & \multicolumn{1}{c|}{\tabincell{c}{Top-5 (\%)}} & Model Type & Design Type & GPU Days \\
\midrule
ResNet-18$^*$~\cite{38}    & 11.7  &   1.8 & 72.5   & -   &    CNN     & Manual           &     -        \\
% MobileNet-V2~\cite{39} &  3.5                      & -                          & 72.0                       & -                        &    CNN     & Manual           &      -       \\
MobileNet-V3~\cite{40} &  5.5                      & -                          &       75.2                 &          -               &    CNN     & Manual           &      -       \\
Deit-Ti~\cite{3}     &  5.7                           &  1.2  &       72.2                     &         91.1                    & Transformer & Manual     &       -      \\
TNT-Ti~\cite{TNT}     &  6.1                           &  1.4  &       73.9                     & 91.9                    & Transformer & Manual     &       -      \\
ViT-Ti~\cite{2}     &  5.7                           & -      &       74.5                     & -                             & Transformer & Manual     &      -       \\
CPVT-Ti~\cite{36} & 6.0                           & -      & 74.9                     & 92.6                                & Transformer & Manual     &      -       \\
PVT-Tiny~\cite{33} &  13.2                           & 1.9      &       75.1                     & -                                & Transformer & Manual     &      -       \\
ViTAS-C~\cite{8}    &  5.6                           & 1.3  &       74.7                     &         91.6                    & Transformer & Auto       & 32 \\
AutoFormer-Ti~\cite{7}&  5.7                           & 1.3  &       74.7                     &         92.6                    & Transformer & Auto       & 24       \\
GLiT-Ti~\cite{28}   &  7.2                           & 1.4  &       76.3                     &         -                       & Hybrid & Auto       & N/A \\
\rowcolor{gray!30} TF-TAS-Ti (Ours)    & \textbf{5.9}                           &  \textbf{1.4}  &       \textbf{75.3}                     &         \textbf{92.8}                    & Transformer & Auto       &      \textbf{0.5}       \\
\midrule
ResNet-50$^*$~\cite{38}    &  25.6                          & 4.1  & 80.2                     & -         &    CNN     & Manual     &     -        \\
% ResNet50D~\cite{}   &  25.6                          &       &       80.5                     &           -                     &    CNN     & Manual     &      -       \\
RegNetY-4GF~\cite{41} & 20.6                           & -      & 79.4                               & -                                &    CNN     & Manual     &     -        \\
% ResNeSt-50~\cite{42}   & 27.5                           & -      &        81.1                    &             -                   &    CNN     & Manual     &      -       \\
DeiT-S~\cite{3}      & 22.1                           & 4.7  &        79.9                    &          95.0                   &Transformer & Manual     &      -       \\
ViT-S/16~\cite{2}    & 22.1                           & 4.7  &        78.8                    &           -                     &Transformer & Manual     &      -       \\
% T2T-ViT-19~\cite{43}  & 39.0                           & 8.0  &        81.2                    &           -                     &Transformer & Manual     &      -       \\
PVT-Small~\cite{33} &  24.5                           & 3.8      &       79.8                     & -                                & Transformer & Manual     &      -       \\
% DeepViT-S~\cite{37} &  27                           & -      & 81.4                     & -                                & Transformer & Manual     &      -       \\
Swin-T~\cite{SwinTransformer}      & 29.0                & 4.5    & 81.3                     & -              & Transformer & Manual     & -  \\
TNT-S~\cite{TNT}      & 23.8                & 5.2    & 81.5                     & 95.7              & Transformer & Manual     & -  \\
CPVT-S~\cite{36}      & 23.0                & -    & 81.5                     & 95.7              & Transformer & Manual     & -  \\
T2T-ViT\_t-14~\cite{4}      & 21.5                & -    & 81.7                     & -              & Transformer & Manual     & -  \\
ViTAS-F~\cite{8}     &  27.6                          & 6.0  &       80.5                     &          95.1                   &Transformer &Auto        & 32 \\
AutoFormer-S~\cite{7}&  22.9                          &  5.1  &       81.7                     &           95.7                  &Transformer &Auto        & 24       \\
GLiT-S~\cite{28}    &  24.6                          &  4.4  &       80.5                     &            -                    & Hybrid &Auto        & N/A \\
\rowcolor{gray!30} TF-TAS-S (Ours)     & \textbf{22.8}                           &  \textbf{5.0}  &       \textbf{81.9}                    &             \textbf{95.8}                &Transformer & Auto      & \textbf{0.5}            \\
\midrule
ResNet-152$^*$~\cite{38}   & 60.2                             & 11.5    & 81.9                      & -                  &    CNN     & Manual     & -            \\
RegNetY-16GF~\cite{21}& 83.6                             & 15.9     & 80.4                      & -                  &    CNN     & Manual     & -           \\
% ResNeXt-101~\cite{}  & 84                             & 15.6     & 81.5                      &              -                  &    CNN     & Manual     & -            \\
% ResNeSt-101~\cite{42}  & 45                             & -     &      81.6                      &              -                  &    CNN     & Manual     & -            \\
% ViT-B/16~\cite{2}   & 86                             & 18    &      79.7                      &              -                  &Transformer & Manual     & -            \\
PVT-Large~\cite{33} &  61.0                           & 9.8      &       81.7                     & -                                & Transformer & Manual     & -  \\
DeiT-B~\cite{3}      & 86.0                             & 18.0    &       81.8                     &             95.6                &Transformer & Manual     & -  \\
% DeepViT-L~\cite{37}      & 55                & -    & 82.2                     & -              & Transformer & Manual     & -  \\
CPVT-B~\cite{36}      & 88.0                & -    & 82.3                     & -              & Transformer & Manual     & -  \\
TNT-B~\cite{TNT}      & 65.5                & 14.1    & 82.9                     & 96.3              & Transformer & Manual     & -  \\
% Swin-S~\cite{SwinTransformer}      & 50                & 8.7    & 83.0                     & -              & Transformer & Manual     & -  \\
Swin-B~\cite{SwinTransformer}      & 88.0                & 15.4    & 83.5                     & -              & Transformer & Manual     & -  \\
T2T-ViT-24~\cite{4}      & 64.1                & -    & 82.6                     & -              & Transformer & Manual     & -  \\
GLiT-B~\cite{28}      & 96.0                             & 17.0    &       82.3                     & -                                & Hybrid & Auto       & N/A \\
AutoFormer-B~\cite{7}& 54.0                            & 11.0    &  82.4                          &            95.7                 &Transformer & Auto       & 24      \\
\rowcolor{gray!30} TF-TAS-B (Ours)    & \textbf{54.0}                             & \textbf{12.0}    &  \textbf{82.2}                          & \textbf{95.6}                                & Transformer & Auto      & \textbf{0.5}     \\
\bottomrule
\end{tabular}
}
\vspace{-2mm}
\end{table*}

Given a specified parameter constraint, we first randomly sample $8,000$ subnets on one ViT search space. Then, the synaptic diversity score of MSAs and the saliency score of MLPs are calculated as the evaluation rank of each subnet.
Based on the calculated DSS-indicator scores of each ViT architecture, we pick the networks with the highest proxy value as the optimal one.
Finally, we retrain the searched optimal network to obtain its final test accuracy.

\section{Experiments}
\label{sec:Experiments}

\subsection{Implementation Details}
TF-TAS includes a search stage and a re-train stage.
In the search stage, the number of sub-networks randomly sampled from the given ViT search space is set to $8,000$, whose weights are randomly initialized. Their proxy scores are calculated for each sub-network.
To compute the proposed DSS-indicator, the input is constructed with each pixel being 1.
After the DSS-indicator of each sub-network is computed, we retrain the top-1 sub-network.   
In the retrain stage, we follow the training configuration in AutoFormer~\cite{7} to train the obtained optimal ViT networks: AdamW optimizer~\cite{loshchilov2018AdamW} with weight decay $0.05$, initial learning rate $1 \times 10^{-3}$ and minimal learning rate $1 \times 10^{-5}$ with cosine scheduler, $5$ epochs warmup, batch size of $256$, and the models are trained with $300$ epochs, \emph{etc.}
All experiments are implemented on NVIDIA Tesla V100 GPUs and the results are estimated on ImageNet~\cite{26}, CIFAR-10/CIFAR-100~\cite{cifar}, and the COCO 2017 dataset~\cite{coco}. 
The image resolution is $224 \times 224$ by default. We also use MindSpore to validate the generalizability of our method. Please refer to the supplementary for the pseudocode of DSS-indicator.

\subsection{Results on AutoFormer Search Space.}
We first evaluate TF-TAS on the search space of AutoFormer, \emph{i.e.}, AutoFormer search space $\mathcal{S}_{A}$.
We compare the performance of searched optimal ViTs with that of \textit{state-of-the-art} TAS methods~\cite{8,28,7} and manually designed CNNs and ViTs~\cite{3,2,TNT,40,SwinTransformer} on ImageNet.

\begin{table}[h]
\centering
\caption{The comparison results on the PiT search space. $\dagger$ indicates the results we reproduce.%The image resolution is $224 \times 224$.
}
\label{tab:OnTheHybridSearchSpace}
\footnotesize{
\setlength{\tabcolsep}{1.1mm}
{\begin{tabular}{l|cccc}
\toprule
Models       & \#Param (M) & FLOPs (B) & Top-1 (\%) & Top-5 (\%) \\
\midrule
PiT-Ti$^{\dagger}$~\cite{17}    & \multicolumn{1}{c}{4.9}        & 0.7      & \multicolumn{1}{c}{73.8}       & \multicolumn{1}{c}{91.7}                \\
PiT-Ti$_{rand}$     & \multicolumn{1}{c}{4.9}        & 0.7      & \multicolumn{1}{c}{69.7}       & \multicolumn{1}{c}{89.1}               \\
\rowcolor{gray!30} TF-TAS-Ti (Ours)    &\multicolumn{1}{c}{\textbf{4.6}}         & \textbf{0.6}      & \multicolumn{1}{c}{\textbf{73.7}}       & \multicolumn{1}{c}{\textbf{91.7}}                  \\
\midrule
PiT-XS$^{\dagger}$~\cite{17}    & \multicolumn{1}{c}{10.6}       & 1.4      & \multicolumn{1}{c}{78.2}       & \multicolumn{1}{c}{94.0}                \\
PiT-XS$_{rand}$     & \multicolumn{1}{c}{10.5}       & 1.8      & \multicolumn{1}{c}{74.8}       & \multicolumn{1}{c}{92.2}                      \\
\rowcolor{gray!30} TF-TAS-XS (Ours)    & \multicolumn{1}{c}{\textbf{10.0}}       & \textbf{1.8}     & \multicolumn{1}{c}{\textbf{77.7}}       & \multicolumn{1}{c}{\textbf{93.8}}                       \\
\midrule
PiT-S$^{\dagger}$~\cite{17} & \multicolumn{1}{c}{23.5}       & 2.9      & \multicolumn{1}{c}{79.9}       & \multicolumn{1}{c}{94.4}                \\
PiT-S$_{rand}$      & \multicolumn{1}{c}{24.2}       & 3.3      & \multicolumn{1}{c}{75.1}       & \multicolumn{1}{c}{92.4}                      \\
\rowcolor{gray!30} TF-TAS-S (Ours)     & \multicolumn{1}{c}{\textbf{23.8}}       & \textbf{3.2}      & \multicolumn{1}{c}{\textbf{80.5}}       & \multicolumn{1}{c}{\textbf{94.9}}                      \\
\midrule
\multirow{2}{*}{Backbone} & \#Param & \multicolumn{3}{c}{Avg. Precision at IOU} \\
  & (M) & AP & AP$_{50}$ & AP$_{75}$ \\
\midrule
ResNet-50~\cite{38} & 41.0 & 41.5 & 60.5 & 44.3 \\
ViT-S~\cite{2} & 34.9 & 36.9 & 57.0 & 38.0 \\
\rowcolor{gray!30} \textbf{Ours} & \textbf{36.0} & \textbf{39.7} & \textbf{60.9} & \textbf{40.4} \\
\bottomrule
\end{tabular}
}
\vspace{-2mm}
}
\end{table}

As listed in \cref{tab:MainResultsOnImageNet}, the searched optimal architectures TF-TAS (\ie, TF-TAS-Ti, TF-TAS-S, and TF-TAS-B) outperforms manually designed CNNs (\eg, ResNet~\cite{38}, MobileNet~\cite{40}, PVT~\cite{33}, and T2T-ViT~\cite{4}) with a clear margin in all the three common model sizes (\ie, tiny, small, and base). Compared with other manually designed ViT architectures \cite{3,2,TNT,36,SwinTransformer}, ours TF-TAS achieves competitive results. Specifically, the searched TF-TAS achieves a top-1 accuracy of 75.3\%, which surpasses DeiT-tiny by 3.1 percent.
Compared with other TAS approaches~\cite{8,7,28} that require larger than $24$ GPU days to seek for optimal ViT architectures, the proposed DSS-indicator helps us achieves the comparable results with much fewer GPU Days. Furthermore, our DSS-indicator comprehensively considers performance and efficiency in searching ViT architectures. Based on the estimation results of the proposed DSS-indicator for each input ViT architecture, we reduce lots of computation budgets in performance estimation and obtain optimal ViT networks with comparable performance within $0.5$ GPU days.
For the analysis of the searched optimal architectures, please refer to the supplementary.

\begin{table}
    \centering
    \caption{Evaluation results of the proposed zero-cost proxy on the state-of-the-art ViT architectures.}
    \label{tab:OnTheSOTAs}
    \scriptsize{
    \setlength{\tabcolsep}{2.8mm}{
    \begin{tabular}{l|cc|c}
        \toprule
        Models & \#Param (M) & Top-1 (\%) & Proxy \\
        \midrule
        % \multicolumn{4}{c}{AutoFormer~\cite{7}} \\
        % \midrule
        % AutoFormer-Ti & 5.7 & 74.7 & $9.6 \times 10^4$ \\
        % AutoFormer-S & 22.9 & 81.7 & $6.2 \times 10^5$ \\
        % AutoFormer-B & 54.0 & 82.4 & $1.9 \times 10^6$ \\
        % \midrule
        \multicolumn{4}{c}{PiT~\cite{17}} \\
        \midrule
        PiT-Ti & 4.9 & 73.8 & $2.9 \times 10^4$ \\
        PiT-XS & 10.6 & 78.2 & $3.4 \times 10^4$ \\
        PiT-S & 23.5 & 80.5 & $4.3 \times 10^4$ \\
        \midrule
        \multicolumn{4}{c}{T2T-ViT~\cite{4}} \\
        \midrule
        T2T-ViT-7 & 4.3 & 71.7 & $1.1 \times 10^5$ \\
        T2T-ViT-10 & 5.9 & 75.2 & $1.3 \times 10^5$ \\
        T2T-ViT-12 & 6.9 & 76.5 & $1.6 \times 10^5$ \\
        T2T-ViT-14 & 21.5 & 81.5 & $5.8 \times 10^5$ \\
        T2T-ViT-19 & 39.2 & 81.9 & $1.1 \times 10^6$ \\
        T2T-ViT-24 & 64.1 & 82.3 & $2.0 \times 10^6$ \\
        % \midrule
        % ViL-X &  &  &  \\
        % ViL-X &  &  &  \\
        % ViL-X &  &  &  \\
        \midrule
        \multicolumn{4}{c}{XCiT~\cite{xcit}} \\
        \midrule
        XCiT-tiny-24-p16 & 12.0 & 79.4 & $3.4 \times 10^4$ \\
        XCiT-small-24-p16 & 48.0 & 82.6 & $9.2 \times 10^4$ \\
        XCiT-medium-24-p16 & 84.0 & 82.7 & $1.4 \times 10^5$ \\
        XCiT-large-24-p16 & 189.0 & 82.9 & $2.4 \times 10^5$ \\
        % \midrule
        % HR-Former-X &  &  &  \\
        % HR-Former-X &  &  &  \\
        % HR-Former-X &  &  &  \\
        \bottomrule
    \end{tabular}
    }
    \vspace{-2mm}
    }
\end{table}

\subsection{Results on PiT Search Space.}
To further investigate the versatility of our DSS-indicator, we build another search space: PiT search space $\mathcal{S}_{P}$.
Without loss of generality, we propose $\mathcal{S}_{P}$ on PiT~\cite{17} and include several important dimensions of ViT (\emph{e.g.,} depth, head number of MSA, MLP ratio), together with depth-wise convolution operations. For the detailed information about $\mathcal{S}_{P}$, please refer to the supplementary.

As listed in \cref{tab:OnTheHybridSearchSpace}, within the budget of $0.5$ GPU days, the proposed DSS-indicator is still able to obtain the optimal ViT architectures that have the comparable or even better Top-1 classification accuracy to PiT-Ti and PiT-S.
The searched networks outperform the randomly selected ones, PiT-Ti$_{rand}$ and PiT-S$_{rand}$, by about $2.9 \sim 5\%$.
We further conduct the transfer experiment on detection with COCO 2017 dataset~\cite{coco} using a simplified setting~\cite{17}. As shown in Tab.\ref{tab:OnTheHybridSearchSpace}, TF-TAS achieves competitive performance in detection task.
These results help us ensure the generalization of the proposed DSS-indicator across different ViT search spaces. 
We also notice that the searched results of TF-TAS on PiT search space are lower than that on AutoFormer search space in \cref{tab:MainResultsOnImageNet}. This observation implies that the search space is also an important part of TAS.

\subsection{On Evaluating Popular Architectures.}
To further investigate the effectiveness and versatility of the proposed proxy, we also conduct evaluation experiments on other popular \textit{state-of-the-art} ViT architectures~\cite{17,4,xcit}. As shown in \cref{tab:OnTheSOTAs}, our DSS-indicator can evaluate the right rank of ViT architectures in their corresponding search space. Interestingly, the values of the proposed proxy obtained across different search spaces are not comparable. This might be caused by several factors. For example, the different ways of model initialization, and the search space itself contains several different modules which makes it difficult to achieve a fair comparison.

\subsection{Transfer Learning Results.}
To test the transferability of the searched optimal ViT networks, we conduct some transfer learning experiments.
We follow the same settings as DeiT~\cite{3} and finetune the TF-TAS-S (see \cref{tab:MainResultsOnImageNet}) in CIFAR-10 (C-10) and CIFAR-100 (C-100)~\cite{cifar}. The results are listed in \cref{tab:DownStreamFineTune}.
As we observe that the optimal ViT architectures found by DSS-indicator in a training-free manner have a similar fine-tuning performance as that of the networks searched by AutoFormer~\cite{7}.

\begin{table}
    \centering
    \caption{The results (\%) on downstream classification datasets. $\uparrow 384$ means that the model is fine-tuned with $384 \times 384$ resolution.}
    \label{tab:DownStreamFineTune}
    \vspace{-1mm}
    \small{
    \setlength{\tabcolsep}{1.2mm}{
    \begin{tabular}{l|c|c|cc}
        \toprule
        Models & \#Param & ImageNet & C-10 & C-100 \\
        \midrule
        ViT-B/16~\cite{2} & 86M & 77.9 & 98.1 & 87.1 \\
        DeiT-B~\cite{3}$\uparrow 384$ & 86M & 83.1 & 99.1 & 90.8 \\
        AutoFormer-S~\cite{7}$\uparrow 384$ & 23M & 83.4 & 99.1 & 91.1 \\
        \midrule
        \textbf{TF-TAS-S$\uparrow 384$ (Ours)} & \textbf{23M} & \textbf{83.5} & \textbf{99.1} & \textbf{91.2} \\
        \bottomrule
    \end{tabular}
    }
    }
    \vspace{-1mm}
\end{table}

\begin{table}
    \centering
    \caption{The Kendall $\tau$ values between various evaluation metrics and the final classification accuracy on the inherit networks randomly sampled from three pre-trained AutoFormer supernets.}
    \label{tab:OnTheAutoFormerBenchmark}
    \vspace{-1mm}
    \small
    {
    \setlength{\tabcolsep}{3.2mm}{
    \begin{tabular}{l|ccc}
        \toprule
        \multirow{2}{*}{Proxy} & \multicolumn{3}{c}{\#Param (M)} \\
        \cline{2-4}
         & 5 - 7 & 15 - 19 & 23 - 25\\
        \midrule
        SNIP~\cite{10}      & 0.481 & 0.028 & -0.282 \\
        GraSP~\cite{11}     & 0.053 & -0.022 & -0.029 \\
        TE-score~\cite{16}  & -0.039 & -0.248 & -0.075 \\
        NASWOT~\cite{15}    & 0.378 & 0.171 & 0.208  \\
        
        \midrule
        \textbf{DSS-indicator (Ours)} & \textbf{0.697} & \textbf{0.615} & \textbf{0.306} \\
        \bottomrule
    \end{tabular}
    }
    \vspace{-2mm}
    }
\end{table}

\subsection{Comparison of Zero-Cost Proxies.}
\label{sec:VS_other_zero-cost}
For full investigation, we compare our DSS-indicator with alternative \textit{state-of-the-art} zero-cost proxies~\cite{10,11,16,15} on CNN search spaces.
To build a reliable test-bed to evaluate these zero-cost proxies, we need a ViT benchmark and we resort to AutoFormer.
We call the search space of AutoFormer~\cite{7} as $\mathcal{S}_{A}$ for simplicity. Empirically, Chen~\etal~\cite{7} find that, the subnet from $\mathcal{S}_{A}$ with its weights inherited from the pre-trained supernet can achieve the performance comparable to that of the retrained one.
Building on this observation, we sample $3,000$ subsets from $\mathcal{S}_{A}$ and obtain their accuracies when they inherit their weights from the pre-trained supernet.
Without loss of generality, we sample the subnet with the amount of parameter in three common ranges: $5$M $\sim 7$M, $15$M $\sim 19$M, and $23$M $\sim 25$M.
With this ViT benchmark, we compare our DSS-indicator with four cutting-eage zero-cost proxy methods: SNIP~\cite{10}, GraSP~\cite{11}, NASWOT~\cite{15} and TE-score~\cite{16}.

\begin{figure*}
    \centering
    \begin{subfigure}[b]{0.329\textwidth}
         \centering
         \includegraphics[width=\textwidth,height=3.85cm]{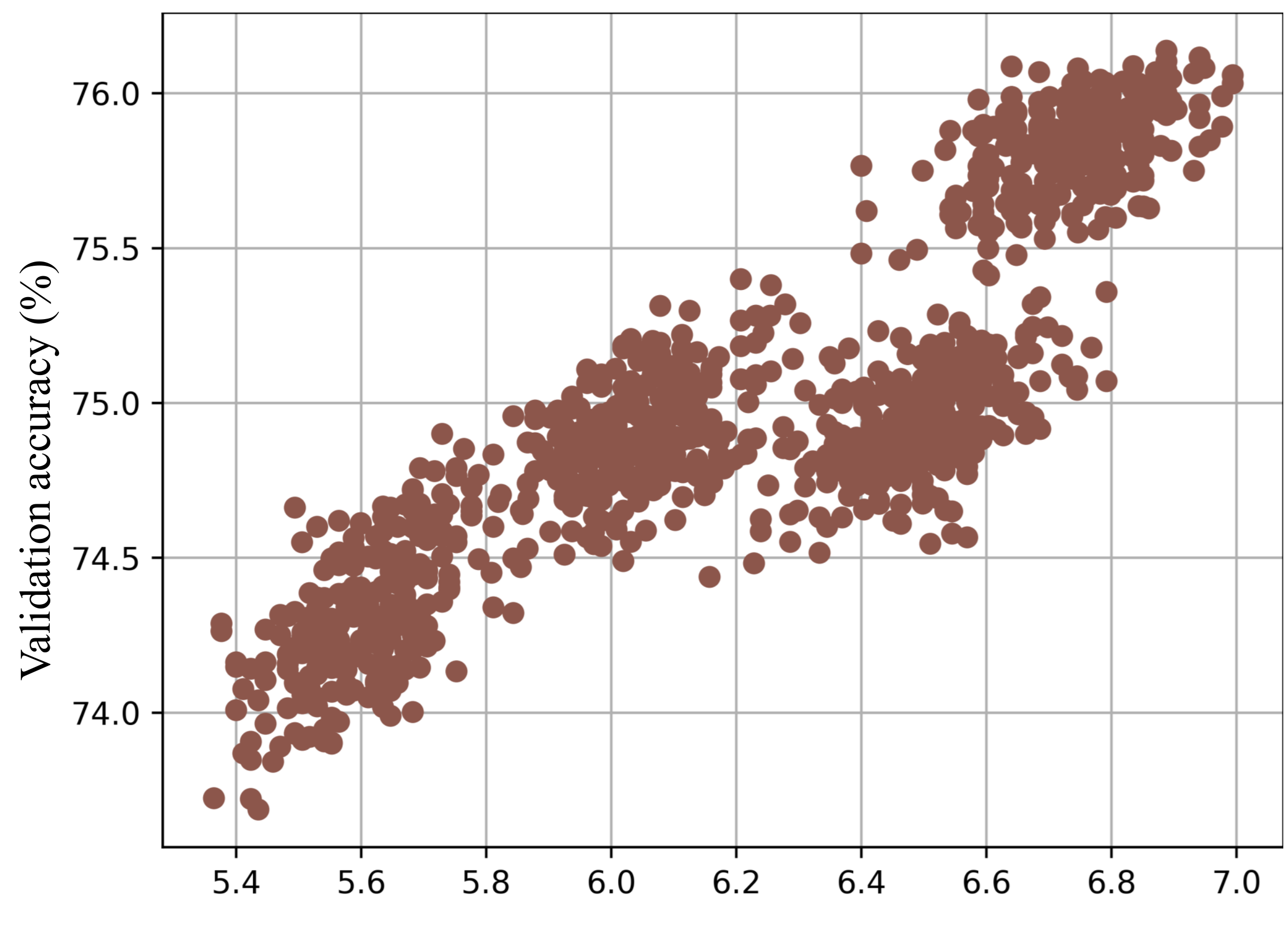}
         \caption{The distribution of the sampled networks.}
         \label{distribution}
    \end{subfigure}
    % \hfill
    \begin{subfigure}[b]{0.329\textwidth}
         \centering
         \includegraphics[width=\textwidth,height=3.85cm]{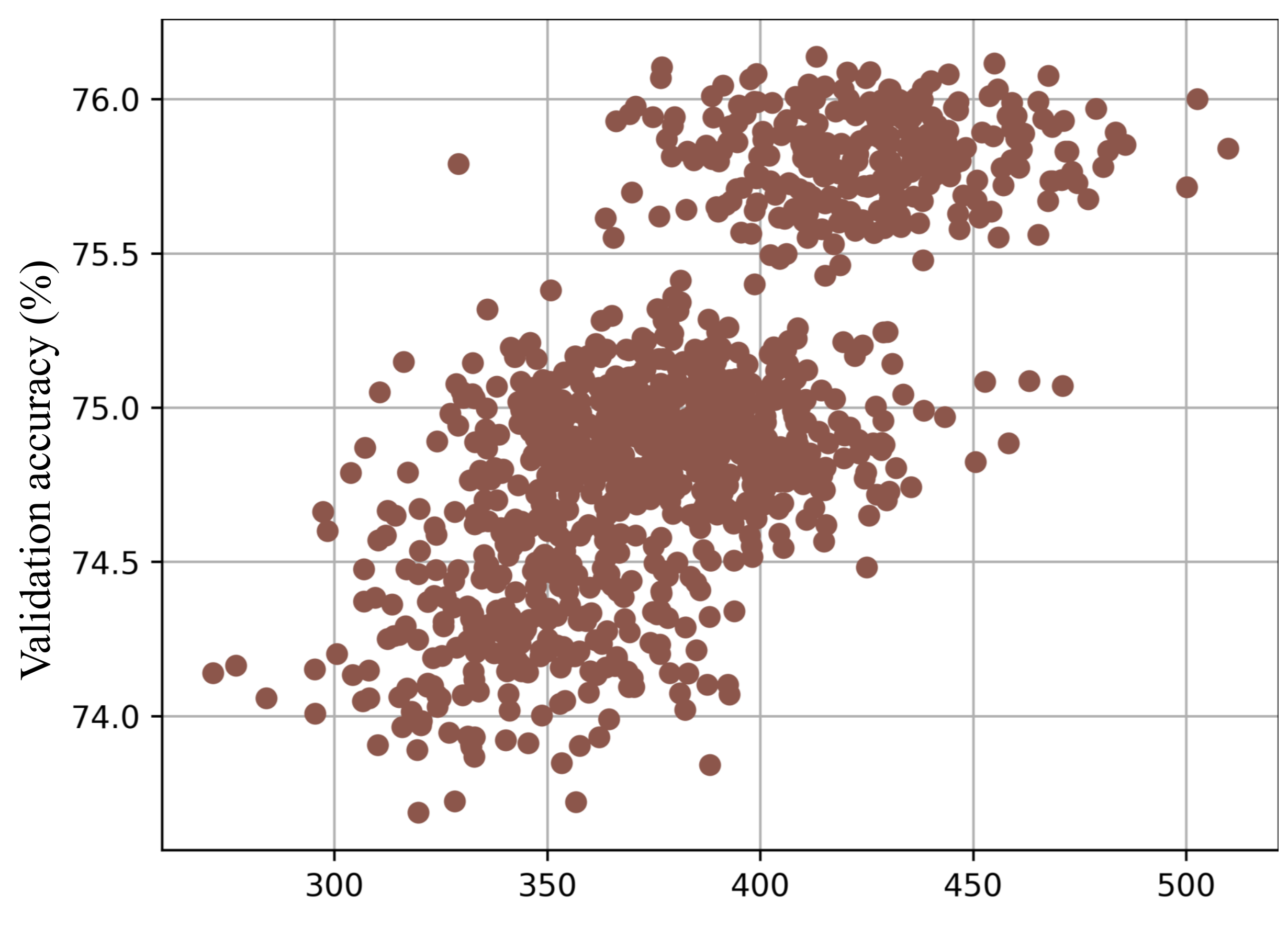}
         \caption{SNIP ($\tau$ = 0.481)}
         \label{SNIP_211115}
    \end{subfigure}
    % \hfill
    \begin{subfigure}[b]{0.329\textwidth}
         \centering
         \includegraphics[width=\textwidth,height=3.85cm]{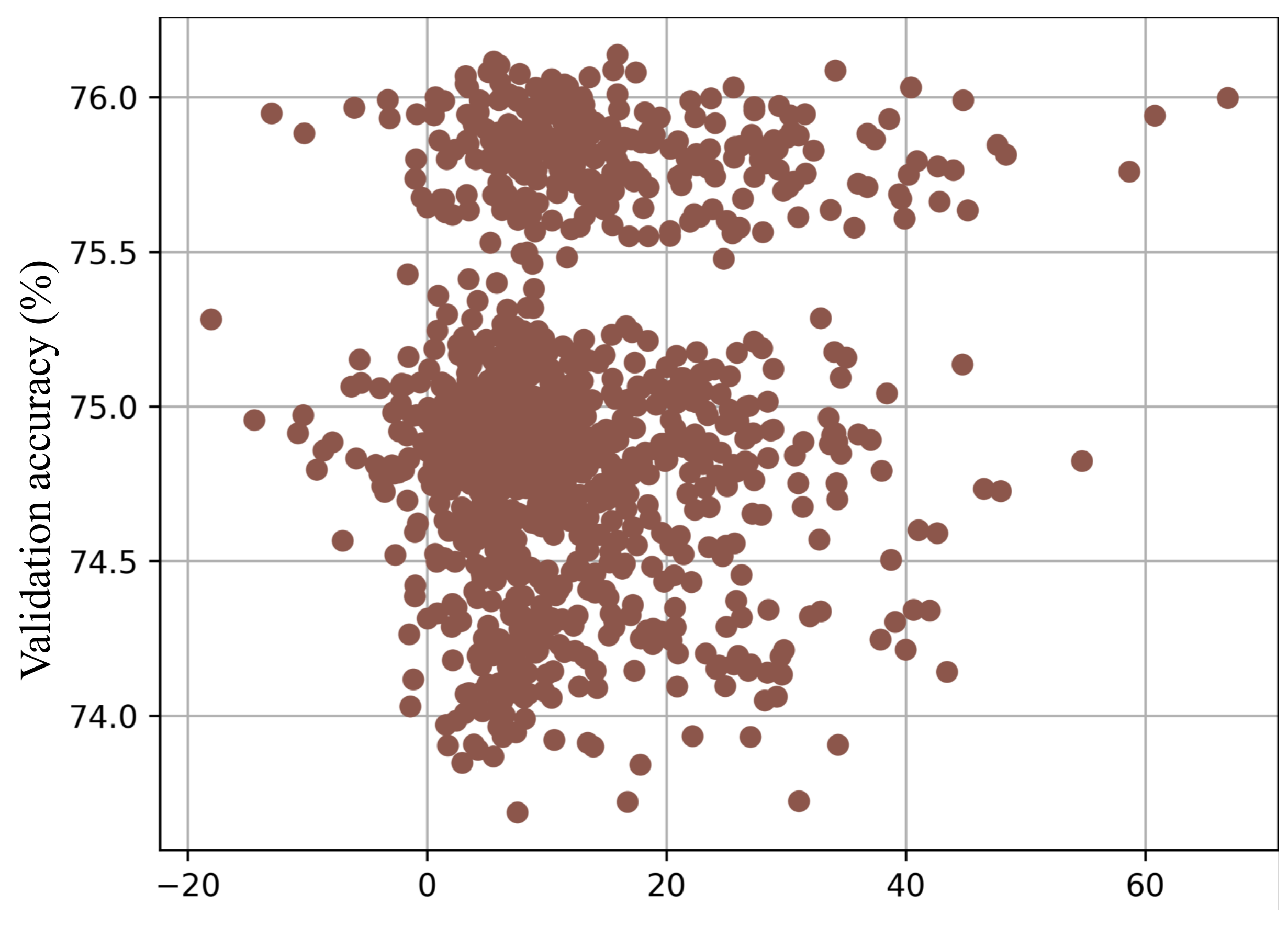}
         \caption{GraSP ($\tau$ = 0.053)}
         \label{GraSP_211115}
    \end{subfigure}
    
    \begin{subfigure}[b]{0.329\textwidth}
         \centering
         \includegraphics[width=\textwidth,height=3.85cm]{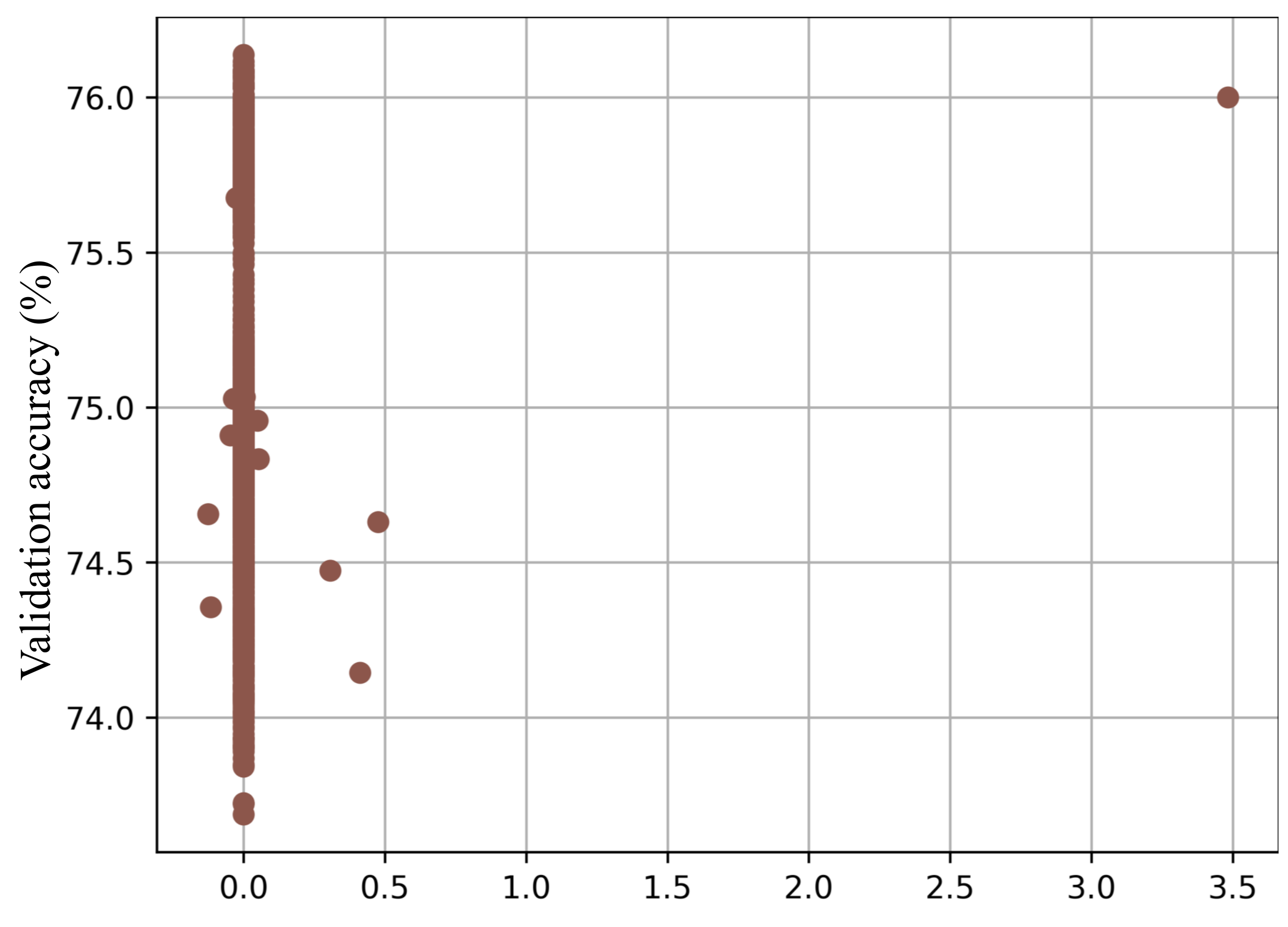}
         \caption{TE-score ($\tau$ = -0.039)}
         \label{NTK_211115}
    \end{subfigure}
    % \hfill
    \begin{subfigure}[b]{0.329\textwidth}
         \centering
         \includegraphics[width=\textwidth,height=3.85cm]{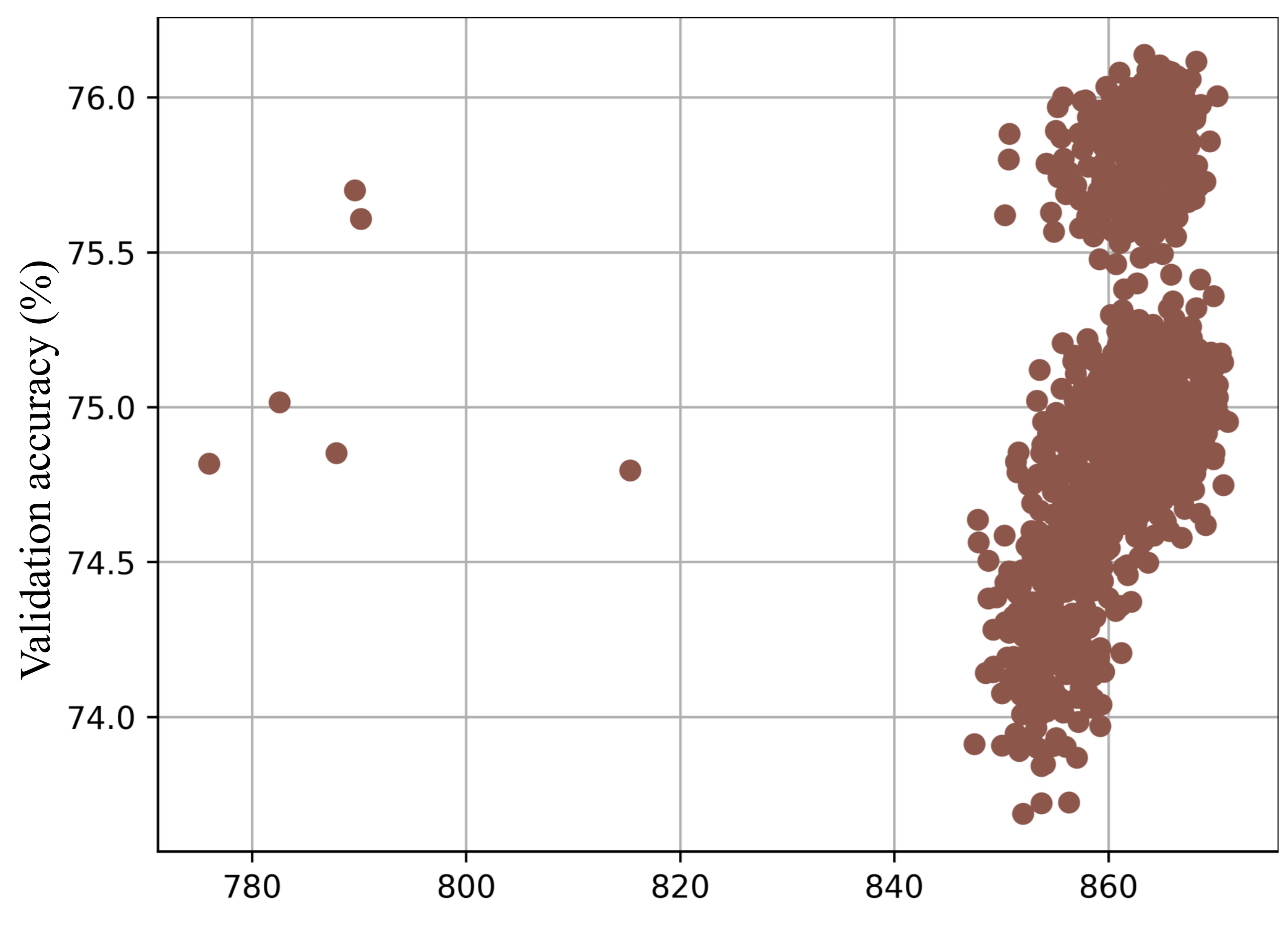}
         \caption{NASWOT ($\tau$ = 0.378)}
         \label{JacobCov_1}
    \end{subfigure}
    % \hfill
    \begin{subfigure}[b]{0.329\textwidth}
         \centering
         \includegraphics[width=\textwidth,height=3.85cm]{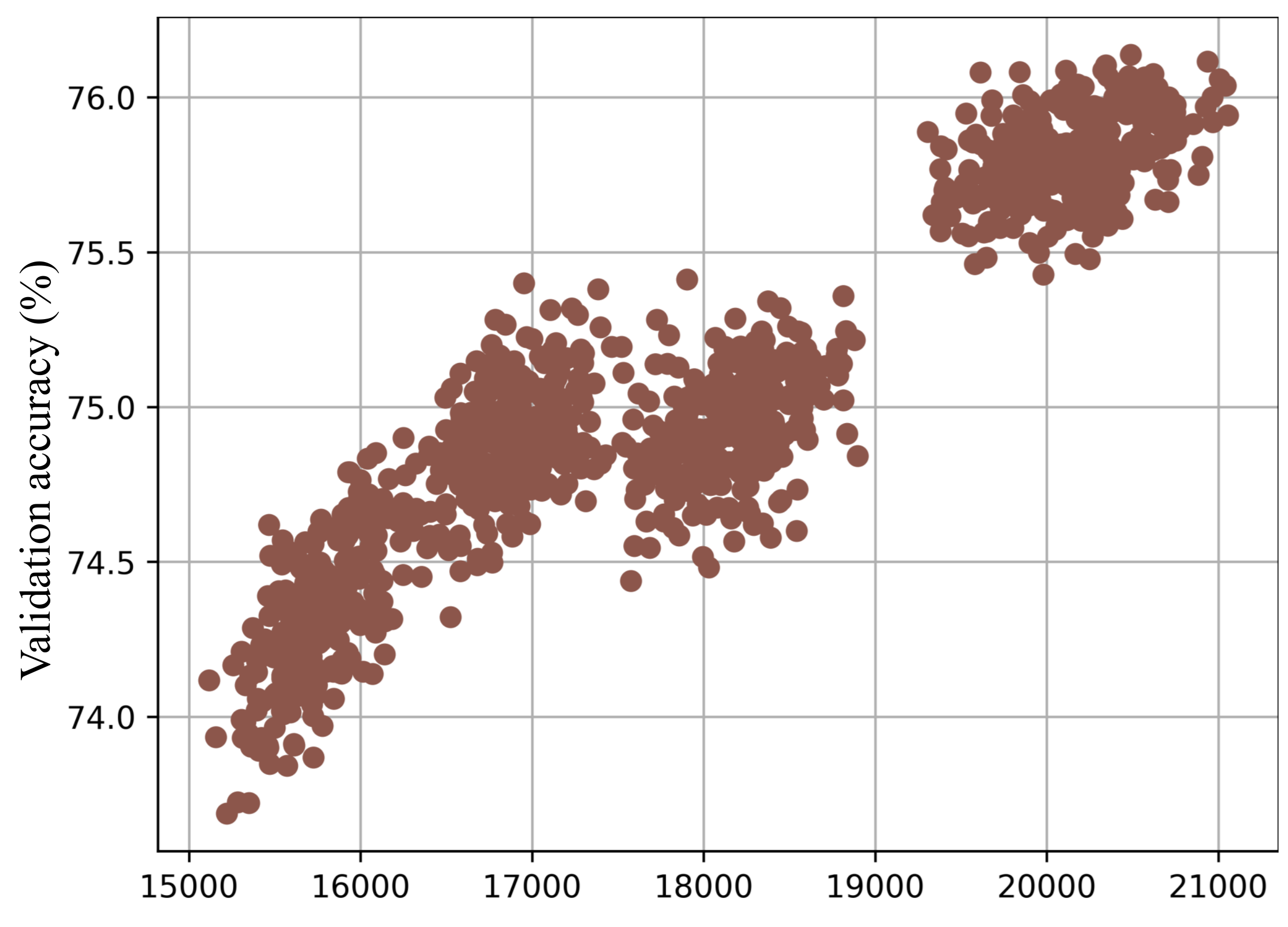}
         \caption{The proposed DSS-indicator ($\tau$ = 0.697)}
         \label{GradNorm_1}
    \end{subfigure}
    \vspace{-5mm}
    \caption{(\textbf{a}): The distribution plot of the $1,000$ subnets (\#param $\in$ [5M, 7M]) sampled from $\mathcal{S}_{A}$. Based on this proxy ViT benchmark, we compare the proposed DSS-indicator with the other counterparts. (\textbf{b}) - (\textbf{f}): The Kendall's $\tau$ rank correlation between the evaluation scores of zero-cost metrics and the classification accuracy on the inherit networks from one pre-trained AutoFormer-Tiny supernet.
    }
    \label{fig:OnTheAutoFormerBenchmark}
    \vspace{-4mm}
\end{figure*}

The results of Kendall $\tau$~\cite{27} are illustrated in \cref{tab:OnTheAutoFormerBenchmark} and \cref{fig:OnTheAutoFormerBenchmark}.
Overall, the relative ranking of the proxies is: Ours $> $ NASWOT $> $ SNIP $> $ GraSP $> $ TE-score.
Our DSS-indicator outperforms the others in ranking various ViT architectures.
The results also provide practical insights to design an effective zero-cost proxy for TAS:
1) Both MSA and MLP should be taken into consideration to rank ViT effectively. And that is the main reason why our DSS-indicator is better than the other alternatives in ranking ViT networks. 2) Based on the results of SNIP~\cite{10} and our DSS-indicator, it is clear that the gradient matrix from the initialized ViT network contains rich information to evaluate the corresponding model. 3) Based on the performance of GraSP~\cite{11} and TE-score~\cite{16}, we find that: despite its practical value in CNNs~\cite{11}, the Hessian matrix of ViT is not easy to use and requires further efforts.

\begin{table}
    \centering
    \caption{The affection of different initialization seeds on the evaluation results of various zero-cost proxies.}
    \label{tab:OnDifferentSeeds}
    \setlength{\tabcolsep}{0.7mm}
    \small{
    \begin{tabular}{l|cccc|cc}
    \toprule
    \multirow{2}{*}{Proxy}      & \multicolumn{4}{c|}{random seed} & \multirow{2}{*}{AVG} & \multirow{2}{*}{STD} \\
    \cline{2-5}
        & 0 & 1 & 2 & 3 & \\
    \midrule
    SNIP~\cite{10}        & 0.481 & 0.530 & 0.486 & 0.507 & 0.501 & 0.019  \\
    GraSP~\cite{11}       & 0.053 & 0.126 & 0.138 & 0.152 & 0.117 & 0.038  \\
    TE-score~\cite{16}    & -0.039 & -0.003 & -0.04 & 0.013 & -0.017 & 0.023  \\
    NASWOT~\cite{15}      & 0.378 & 0.332 & 0.394 & 0.421 & 0.381 & 0.032  \\
    \midrule
    DSS-indicator (Ours)  & 0.697 & 0.697 & 0.697 & 0.697 & 0.697 & 0 \\
    \bottomrule
    \end{tabular}
    }
    \vspace{-4mm}
\end{table}

%%%%%%%%%
\subsection{Consistency across different random seeds.}
To check the stability of various proxies, we generate the results with four random seeds and calculate the statics.
For simplicity, the experiments are conducted when the amount of parameter of ViT network is among $5 \sim 7$M.
As listed in \cref{tab:OnDifferentSeeds}, there are certain fluctuations on several proxies under different seeds. Our DSS-indicator is invariant to different seeds (as mentioned in \cref{Ourmethod}). For different seed influences the sampling of input data, the proxies that use the sampled data as the input could be unstable.

% %%%%%%%%%

\vspace{-1mm}
\section{Conclusion}
To accelerate the searching efficiency of TAS, for the first time, we propose an effective zero-cost proxy in evaluating ViT architectures. In specific, a ViT-oriented performance indicator, \ie, DSS-indicator, is proposed.
It is built on two theoretical perspectives: synaptic diversity and synaptic saliency.
Based on these two dimensions, the proposed indicator measures the synaptic diversity on MSAs and the synaptic saliency on MLPs, respectively. Compared to other cutting-edge TAS methods, the random search guided by our DSS-indicator achieves a competitive performance among different popular ViT search spaces. And significantly, we greatly promote the searching efficiency of TAS: it merely takes $0.5$ GPU days to seek relatively optimal ViT architecture, compared with $24$ GPU days by the existing counterparts.

\vspace{2mm}
\noindent\textbf{Acknowledgments:}
\small{
This work was supported by the National Science Fund for Distinguished Young Scholars (No.62025603), the National Natural Science Foundation of China (No.U21B2037, No.62176222, No.62176223, No.62176226, No.62072386, No.62072387, No.62072389, and No.62002305), Guangdong Basic and Applied Basic Research Foundation (No.2019B1515120049), the Natural Science Foundation of Fujian Province of China (No.2021J01002), and CAAI-Huawei MindSpore Open Fund.
}

%%%%%%%%% REFERENCES
% \small
{
\bibliographystyle{ieee_fullname}
\bibliography{MAIN}
}

\end{document}